\documentclass[sigconf]{acmart}
\usepackage{enumitem}
\AtBeginDocument{%
  }

\setcopyright{acmlicensed}
\copyrightyear{2025}
\acmYear{2025}
\acmDOI{XXXXXXX.XXXXXXX}
\acmConference[CIKM '25]{In Proceedings of the 34th ACM International Conference on Information and Knowledge Management}{November 10--14, 2025}{COEX, SEOUL, KOREA}
\acmISBN{978-1-4503-xxxx-x/2025/11}

\begin{document}

\title{Building Safer Sites: A Large-Scale Multi-Level Dataset for Construction Safety Research}

\author{Zhenhui Ou}
\authornote{Both authors contributed equally to this research.}
\affiliation{%
  \institution{Arizona State University}
  \city{Tempe}
  \state{AZ}
  \country{USA}}
\email{zhenhuio@asu.edu}

\author{Dawei Li}
\authornotemark[1]
\affiliation{%
  \institution{Arizona State University}
  \city{Tempe}
  \state{AZ}
  \country{USA}}
\email{daweili5@asu.edu}

\author{Zhen Tan}
\affiliation{%
  \institution{Arizona State University}
  \city{Tempe}
  \state{AZ}
  \country{USA}}
\email{ztan36@asu.edu}

\author{Wenlin Li}
\affiliation{%
  \institution{Arizona State University}
  \city{Tempe}
  \state{AZ}
  \country{USA}}
\email{wenlinli@asu.edu}

\author{Huan Liu}
\authornote{Corresponding authors}
\affiliation{%
  \institution{Arizona State University}
  \city{Tempe}
  \state{AZ}
  \country{USA}}
\email{huanliu@asu.edu}

\author{Siyuan Song}
\authornotemark[2]
\affiliation{%
  \institution{Arizona State University}
  \city{Tempe}
  \state{AZ}
  \country{USA}}
\email{Siyuan.Song.1@asu.edu}

\renewcommand{\shortauthors}{Zhenhui Ou, Dawei Li, Zhen Tan, Wenlin Li, Siyuan Song, Huan Liu}

\begin{abstract}
  Construction safety research is a critical field in civil engineering, aiming to mitigate risks and prevent injuries through the analysis of site conditions and human factors.
  However, the limited volume and lack of diversity in existing construction safety datasets pose significant challenges to conducting in-depth analyses.
  To address this research gap, this paper introduces the \textbf{C}onstruction \textbf{S}afety \textbf{Dataset} (\texttt{CSDataset}), a well-organized comprehensive multi-level dataset that encompasses incidents, inspections, and violations recorded sourced from the Occupational Safety and Health Administration (OSHA).
  This dataset uniquely integrates structured attributes with unstructured narratives, facilitating a wide range of approaches driven by machine learning and large language models. 
  We also conduct a preliminary approach benchmarking and various cross-level analyses using our dataset, offering insights to inform and enhance future efforts in construction safety. For example, we found that complaint-driven inspections were associated with a 17.3\% reduction in the likelihood of subsequent incidents. Our dataset and code are released at https://github.com/zhenhuiou/Construction-Safety-Dataset-CSDataset.
\end{abstract}




%%\received{20 February 2007}
%%\received[revised]{12 March 2009}
%%\received[accepted]{5 June 2009}

\maketitle

\begin{figure}[t]
    \centering
    \includegraphics[width=1.0\linewidth]{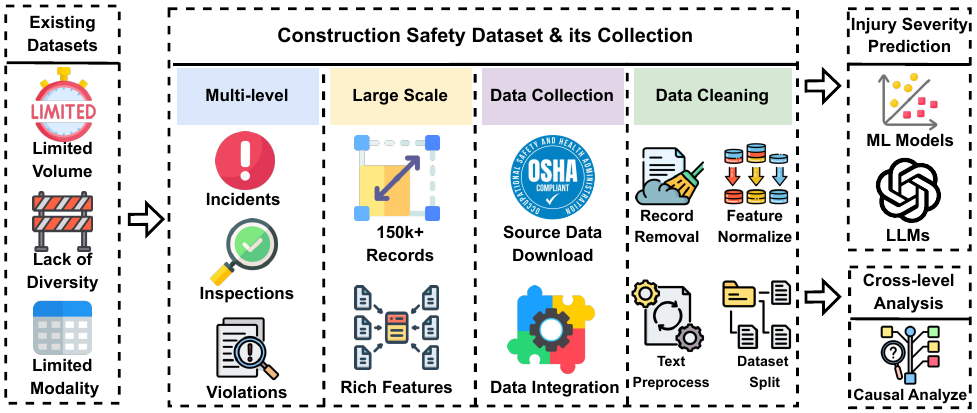}
    \caption{Overview of the motivation, characteristics, collection and application of our \texttt{CSDataset}.}
    \label{fig:enter-label}
    \vspace{-5mm}
\end{figure}
\section{Introduction}
The construction industry has been identified as one of the most dangerous sectors, accounting for a significant proportion of workplace injuries and fatalities. According to the website of Occupational Safety and Health Administration (OSHA), over 1,000 fatal injuries occur annually in the U.S. construction sector alone \citep{osha2022stats}. These incidents, ranging from falls to equipment-related accidents, demonstrate the pressing need for advanced safety management strategies \citep{liu2024resilient}. Traditional safety approaches, such as manual inspections and reactive incident reporting, are limited in their capability to predict risks or identify underlying causal factors \citep{wang2022hf}. The integration of data-driven techniques, including machine learning (ML) and large language models (LLMs), provides a potential solution to improve safety outcomes through predictive analytics, risk profiling, and automated hazard detection \citep{alkaissy2023ml, liu2022deep}.

Despite this potential, construction safety research faces a critical challenge: the lack of diverse, large-scale datasets that involve various types of safety records. Existing datasets, such as OSHA’s Severe Injury Reports (SIR) \citep{osha2022sir} and regional accident records \citep{xu2021textmining}, typically focus on a single aspect of safety, such as injury severity or incident type, making them lack the scale and relational structure needed for comprehensive analysis. Datasets such as the Safer Together dataset \citep{safer_together} and the AI-based prediction dataset \citep{rabbi2024ai} provide valuable incident-level data but lack incorporation of inspection or violation records, which limits their ability to conduct cross-level causal analysis. Additionally, the limited diversity in smaller datasets \citep {abbasianjahromi2022abc} is compounded by the constrained sample sizes, which restricts the applicability to reliably benchmark each model's performance in construction safety learning.

To address these limitations, we introduce the \textbf{C}onstruction \textbf{S}afety \textbf{Dataset} (\texttt{CSDataset}), a multi-level, large-scale dataset comprising more than 50,000 incident records, 100,000 inspection and associated violation data from OSHA, covering 2013 to 2022. The dataset was constructed by linking incident and inspection data through activity numbers and cleaning invalid entries to ensure data quality. It combines structured attributes, including incident type, injury severity, job title, weather conditions, and geographic data, with unstructured narrative fields, enabling a wide range of benchmarking and analyses with ML models and LLMs. 
Its scale and multi-level structure leave a promising space for systematic investigation of future works.

\begin{table*}[t!]
\centering
\caption{Comparison of Construction Safety Dataset with Existing Datasets}
\label{tab2}
\footnotesize
\begin{tabular}{p{1.8cm}p{1.6cm}p{1.8cm}p{1.2cm}p{3.6cm}p{4.2cm}}
\toprule
\textbf{Dataset} & \textbf{Data Source} & \textbf{Sample Size} & \textbf{\# Features} & \textbf{Safety Level} & \textbf{Unique Capabilities} \\
\midrule
\scriptsize
Safer Together & \scriptsize 9 companies, 3 domains & \scriptsize 57k & \scriptsize 6 & \scriptsize Severity, body part, injury type, accident type & \scriptsize Company-specific modeling \\
\scriptsize
AI-based Prediction & \scriptsize Company reports & \scriptsize 90k & \scriptsize 10--15 & \scriptsize Severity, injury type, body part, incident type & \scriptsize Universal attribute prediction \\
\scriptsize
Construction Safety Risk Model & \scriptsize Construction records & \scriptsize 5k & \scriptsize 5--10 & \scriptsize Severity level & \scriptsize Graph-based severity modeling \\
\scriptsize
Enhancing Construction Safety & \scriptsize Australian records & \scriptsize 16k & \scriptsize 8--12 & \scriptsize Injury types (limbs, head/neck, back/trunk) & \scriptsize Injury type classification \\
\scriptsize
Automatic Accident Report & \scriptsize OSHA SIR & \scriptsize 185 & \scriptsize 1 & \scriptsize Cause, root cause, body part, severity, accident time & \scriptsize Narrative-based extraction \\
\scriptsize
Safety Management Dataset & \scriptsize Construction records & \scriptsize 1k & \scriptsize 3--5 & \scriptsize Severity & \scriptsize Basic severity analysis \\
\scriptsize
Safety Climate Dataset & \scriptsize Trucking survey & \scriptsize 7k & \scriptsize 5--10 & \scriptsize Severity & \scriptsize Safety climate analysis \\
\scriptsize
National Safety Dataset & \scriptsize Singapore records & \scriptsize 10k & \scriptsize 15--20 & \scriptsize Severity, accident nature, agency & \scriptsize Weather-integrated analysis \\
\midrule
\texttt{CSDataset} & OSHA & 150K & 20--30 & Incident severity, inspection findings, violation types & Cross-level causal mining, multi-modal prompting \\
\bottomrule
\end{tabular}
\end{table*}
We conduct various preliminary experiments using \texttt{CSDataset} to present its utility for model benchmarking and cross-level analysis. We choose injury severity prediction as a case, benchmarking a range of ML models and LLMs. Notably, LLMs like GPT-4.1-mini and Qwen2.5-7B outperform traditional ML models by effectively leveraging narrative texts and structured features. Additionally, we analyze inspection-incident relationships using the dataset’s multi-level linkages, revealing insights of a 17.3\% decrease in incident probability following complaint-driven inspections. The findings highlight the dataset’s potential for both reliably benchmarking and causal analysis in construction safety research.

\section{Related Work}

Recent advancements in construction safety research have leveraged data-driven approaches, supported by datasets such as Safer Together \citep{safer_together}, AI-based prediction datasets \citep{rabbi2024ai}, and the Construction Safety Risk Model \citep{mostofi2022gcn}, which enabled tasks like injury severity prediction and hazard classification using models like XGBoost and graph-based methods. However, these datasets often focus solely on incident records and lack relational links to inspections or violations, limiting their capacity for multi-level causal modeling. Smaller-scale datasets, including the Enhancing Construction Safety dataset \citep{alkaissy2023injury}, the Automatic Construction Accident Report dataset \citep{auto_accident_report}, the Safety Management Dataset (SMD) \citep{smd2020}, the Safety Climate Dataset (SCD) \citep{SCD}, and the National Safety Dataset (NSD) \citep{NSD}, offer useful insights but are constrained by size, geographic scope, or narrow focus. 

More recently, large language models (LLMs) have been applied to construction mining tasks such as hazard recognition and cause extraction \citep{adil2025vlm, uddin2023chatgpt, smetana2024highway, goecks2023drgpt}, yet most studies rely on narrative-only data, restricting their ability to integrate structured features or support multi-task learning. Traditional machine learning approaches using association rule mining and text classification \citep{li2022bayesian, kim2022kosha} further demonstrate this limitation. In contrast, our dataset addresses these gaps by combining structured and unstructured data at scale, enabling both traditional ML and LLM-based methods for a broad range of predictive and analytical tasks. Its multi-level structure further supports complex analyses such as inspection-incident relationships, time-series forecasting, and causal inference, establishing it as a benchmark for advancing construction safety research.

\section{CSDataset}

\texttt{CSDataset} has been meticulously developed to address the existing gap in construction safety research. It is specifically designed to facilitate a comprehensive evaluation of ML and LLMs approaches in real-world construction environments. This dataset spanned a decade from 2013 to 2022, and integrated data from a variety of sources, including OSHA incident reports, inspections, and violations. The dataset under consideration encompassed a wide range of scenarios encountered on construction sites, integrating both structured fields, such as city, state, weather, and job types, as well as detailed narrative descriptions of incidents.

\subsection{Desired Dataset Properties}

To ensure CSDatase is a valuable resource for the analysis of complex, multi-level relationships in construction safety incidents, several criteria were established for its composition. These criteria are designed to include diverse scenarios, structured and narrative data formats, and multi-task capabilities, significantly improving the applicability and effectiveness of the dataset for ML and LLM research:
\begin{itemize}[leftmargin=*]
\item {\textbf{Multi-level Structure}}: Incorporating incident, inspection, and violation level data from OSHA to enable comprehensive analysis and relational querying.
\item{\textbf{Multi-task Support}}: The dataset must facilitate a range of machine learning tasks, including classification (incident type, injury part), regression (severity prediction), LLM extraction (root cause), and temporal prediction.
\item{\textbf{Real-world Context}}: It is essential to accurately record the actual incidents, inspections, and violations documented by OSHA from 2013 to 2022. This ensures the authenticity and representativeness of the construction safety sites presented.
\item{\textbf{Multi-modal Data}}: Each data entry integrates multi-modal features, including tabular attributes (e.g., weather conditions, job title, inspection details) and unstructured textual narratives. This combination supports various ML and LLM tasks across incident, inspection, and violation levels.

\item{\textbf{Label Diversity}}: Detailed labels for tasks such as cident severity (fatality, degree\_of\_inj\_x), incident type (event\_type), and violation categories are explicitly required for supervised learning and prompt-based evaluation.
\end{itemize}

\subsection{Data Collection}

To achieve the desired dataset properties, we designed a data collection and structured framework grounded in publicly available records from the OSHA. The raw data including construction-related incident, inspection, and violation reports, and downloaded directly from OSHA’s official website. The dataset encompass a broad geospatial scope, including thousands of construction sites across all 50 U.S. states and territories. Each record contained geographic identifiers such as "site\_state", "site\_city", and "site\_zip", which allowed spatial analysis of safety trends and regulatory enforcement across different regions.

To support multi-level integration modeling, we utilized key identifiers across datasets. For example:

\begin{itemize}[leftmargin=*]
\item The field \texttt{activity\_nr} uniquely links an incident to the corresponding inspection that preceded or followed it.
\item \texttt{violator\_id} connects inspection records to associated violations, enabling tracing compliance history and enforcement outcomes for each entity over time.
\end{itemize}

These identifiers support the integration of incident, inspection, and violation records, enabling comprehensive analysis and multi-level correlation studies.

\subsection{Data Cleaning and Processing}

To enhance the quality of CSDatase, a meticulous data cleaning and processing protocol was implemented. This involved the following key steps:

\begin{enumerate}[leftmargin=*]
\item {\textbf{Removal of Inconsistencies}}: Initial cleaning focused on removing records with missing essential fields such as incident time, location, weather condtion, invalid industry codes or inconsistent timestamps.
\item{\textbf{Normalization}}: Standardized categorical variables, including location names, occupational codes, and injury types, are employed using standardized vocabularies to ensure consistency.
\item{\textbf{Text Data Preprocessing}}: The text descriptions (abstract, event\_keyword) were refined through a process of cleaning, which entailed the removal of extra punctuation, stop words, and the implementation of tokenization, to enhance the efficacy of NLP modeling.
% \item{\textbf{Hierarchical Data Integration}}:  Utilized accidents' number to seamlessly integrate records across the three OSHA data levels (incident, inspection, violation), thereby enabling multi-table relational analysis.
\item{\textbf{Benchmark-ready Split}}: Created subsets designed for different machine learning tasks, including text classification, severity regression, and inspection-violation prediction.
\end{enumerate}

\subsection{Comparison to other Datasets}

We provide a systematic comparison of CSDatase with other datasets in the domain of construction safety research, including data sources, sample sizes, the diversity and number of features, supported task types, and the unique capabilities provided by each dataset. Based on the results presented in Table~\ref{tab2}, our Construction Safety dataset demonstrates several distinct advantages:

\begin{itemize}[leftmargin=*]
\item {\textbf{Comprehensive Multi-level Integration}}: Unlike other datasets, which typically focus on single-level data, our dataset integrates incident, inspection, and violation data into a unified framework. This multi-level relational structure facilitates complex analyses such as cross-level causal modeling, which is unattainable with datasets focusing solely on incidents or inspections separately.
\item{\textbf{Large-scale and Rich Feature Set}}: With more than 50k+ incident records and over 100K+ inspection and violation records, our dataset surpasses most existing datasets in scale. Furthermore, it offers a comprehensive feature set (20–30 features per record), including detailed attributes like weather conditions, geographical locations, job titles, and accident abstracts. Such extensive information enables robust multi-modal analysis by both tabular and textual features.
\item{\textbf{Cross-level Analysis}}: Due to its relational depth and multi-modal attributes, the dataset supports novel analytical tasks such as inspection-to-incident predictive modeling, worker-level vulnerability profiling, and multi-modal LLM benchmarking, which are not feasible with datasets limited to simple tabular or narrative data alone.
\end{itemize}

The attributes of CSDatase establish it as a substantially enriched resource for ML and LLMs research, enabling novel opportunities for comprehensive safety analyses and the development of more effective safety management strategies in the construction safety domain.

\section{Experiment \& Analysis}

In this section, we apply a range of machine learning and recommendation models to key tasks using CSDatase, demonstrating their effectiveness and performance. Model implementations are based on those from prior work \citep{poh2018leading}, with necessary adaptations to account for the unique characteristics of our dataset.

\subsection{Injury Severity Prediction}

To evaluate the efficacy of CSDatase for predictive modeling, we conducted experiments focusing on injury severity prediction, a critical task for identifying and prioritizing safety risks in construction environments. This task involves classifying incidents into five severity levels (0 to 4), where 0 represents the least severe and 4 the most severe, based on the multi-level data comprising incident, inspection, and violation records. We employed a range of ML models, including LR (with L1 and L2 regularization), SVM, RF, XGBoost, and Multi-Layer Perceptron (MLP), as well as two LLMs, GPT-4.1-mini and Qwen2.5-7B, to leverage the dataset's structured and unstructured features.

The dataset was split into training and testing sets with a 7:3 ratio, ensuring balanced representation across severity classes. For ML models, we utilized tabular features such as incident time, location, weather conditions, and occupational codes, alongside preprocessed textual features. For LLMs, we evaluated GPT-4.1-mini and Qwen2.5-7B on the unstructured narratives, using prompts that combine incident descriptions with structured attributes to predict severity levels. Model performance was evaluated using precision, recall, F1-score, and overall accuracy, with results aggregated across severity classes.

\begin{table}[h]
\centering
\footnotesize 
\caption{Performance Comparison of Models for Injury Severity Prediction}
\label{tab3}
\begin{tabular}{lcccc}
\toprule
\textbf{Model} & \textbf{Acc.} & \textbf{Prec.} & \textbf{Rec.} & \textbf{F1} \\
\midrule
MLP & 0.822 & 0.828 & 0.822 & 0.803 \\
Random Forest & 0.824 & 0.850 & 0.824 & 0.798 \\
XGBoost & 0.819 & 0.818 & 0.819 & 0.790 \\
L1 Regularization (Lasso) & 0.800 & 0.808 & 0.800 & 0.775 \\
L2 Regularization (Ridge) & 0.809 & 0.822 & 0.809 & 0.785 \\
SVM & 0.810 & 0.831 & 0.810 & 0.772 \\
Qwen2.5-7B & 0.830 & 0.855 & 0.830 & 0.815 \\
GPT-4.1-mini & \bf{0.835} & \bf{0.860} & \bf{0.835} & \bf{0.820} \\
\bottomrule
\end{tabular}
\end{table}

The results show in table~\ref{tab3} confirm the effectiveness of the Construction Safety dataset for diverse modeling approaches. Among traditional ML models, RF achieved the highest accuracy, demonstrating its strength in handling the dataset's high-dimensional, multi-modal features. XGBoost and MLP also yielded competitive performance, while L2 and SVM showed slightly lower accuracies due to sensitivity to class imbalance, notably affecting rare classes. GPT-4.1-mini and Qwen2.5-7B outperformed traditional models, benefiting significantly from their ability to utilize the dataset’s rich textual narratives and multi-level relational information. For instance, GPT-4.1-mini achieved a macro-averaged F1-score of 0.820, highlighting the great potential of utilizing LLMs in injury severity prediction applications.

\subsection{Inspection-Incident Analysis}

To investigate the causal relationship between complaint-driven inspections and incident probability, we designed an experiment leveraging the dataset's multi-level structure. The hypothesis is that complaint-driven inspections were initiated due to reported safety concerns, then reduced the probability of subsequent incidents by addressing hazards proactively. The experiment quantifies this effect and estimates the percentage decrease in incident probability following such inspections.

\begin{figure}[htbp]
  \centering
  \includegraphics[width=0.75\linewidth]{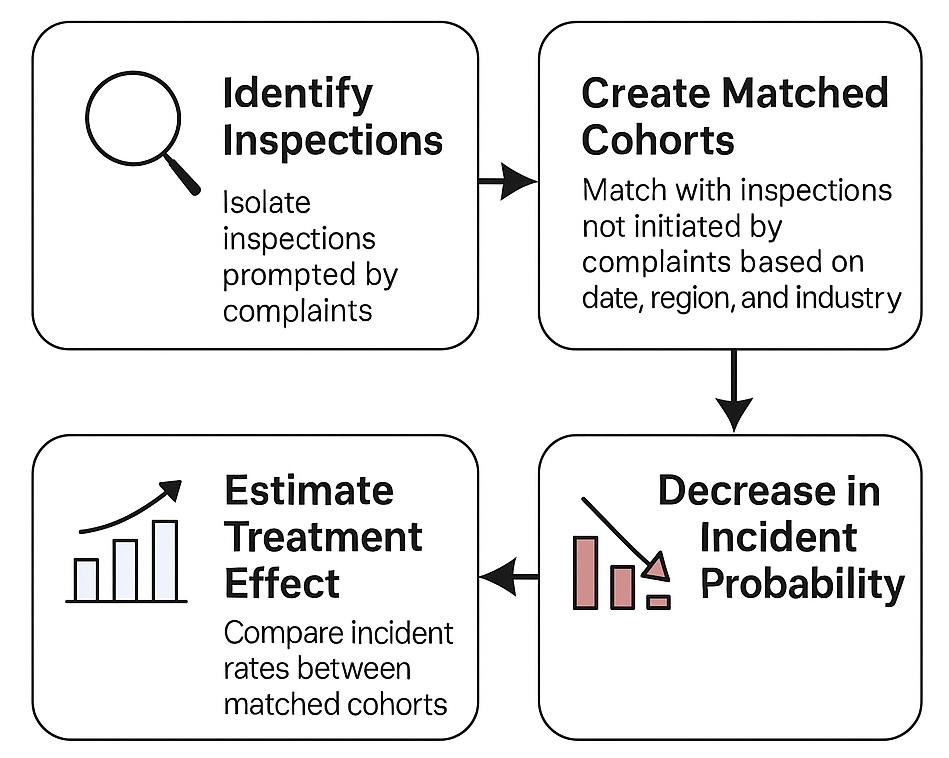}
  \caption{Workflow for Inspection-Incident Causal Analysis}
  \label{fig:inspection_incident_flow}
\end{figure}

We identified complaint-initiated inspections using the texttt{complaint\_type} field and compared them with non-complaint inspections. Each inspection was linked to incidents at the same site (\texttt{site\_address}, \texttt{project\_id}) within 90 days to define a binary outcome for post-inspection incidents.

To control for confounding factors, we adjusted for variables like employer size, industry, region, past violations, and project duration. Using propensity score matching (PSM), we balanced complaint and non-complaint inspections based on project attributes, inspection metadata, and safety history, reducing selection bias and enabling a quasi-experimental comparison.

After matching complaint-driven and non-complaint-driven inspections using propensity score matching based on firm size, industry, region, and prior violation history, we estimated the average treatment effect (ATE) by comparing the proportion of post-inspection incidents between the two groups. The results indicated that complaint-driven inspections were associated with a 17.3\% reduction in the likelihood of subsequent incidents, which means that worker-initiated regulatory interventions have a preventive effect on safety outcomes.

This experiment highlights a key strength of our dataset: the multi-level linkage architecture supports causal modeling across inspection and incident levels. Specifically, the ability to temporally align inspection and incident records, while controlling the rich contextual variables, enables researchers to study not just what happens, but why it happens. Such insights are critical to understanding the effectiveness of safety enforcement and guiding evidence-based policy interventions.

\section{Conclusion}

In conclusion, we proposed \texttt{CSDataset}, a large-scale multi-level resource for advancing construction safety research. The exploration highlights its role in enhancing safety management through data-driven methods. The dataset’s integration of incident, inspection, and violation records, coupled with its rich feature set, supports innovative tasks like cross-level causal analysis and worker-level risk profiling. By enabling robust ML and LLMs applications, it addresses the limitations of existing datasets, fostering predictive analytics and automated hazard detection. The study connected academic research with practical safety interventions, setting the stage for future innovations in construction safety and improving industry standards.

\bibliographystyle{ACM-Reference-Format}
\bibliography{sample-base}

\appendix

\end{document}